\documentclass[conference]{IEEEtran}
\usepackage{times}

\usepackage[numbers]{natbib}
\usepackage{multicol}
\usepackage[bookmarks=true]{hyperref}
\usepackage{graphics} 
\usepackage{epsfig} 
\usepackage{mathptmx} 
\usepackage{times} 
\usepackage{amsmath,bm} 
\usepackage{amssymb}  
\usepackage{booktabs}
\usepackage{bm}
\usepackage{comment}
\usepackage{color}
\usepackage{graphicx}
\usepackage{subcaption}
\usepackage{makecell}
\usepackage{threeparttable}

\renewcommand\theadalign{cl}
\renewcommand\theadfont{\bfseries}
\renewcommand\theadgape{\Gape[4pt]}
\renewcommand\cellgape{\Gape[0pt]}

\newcommand{\bs}[1]{\bm{#1}}
\newcommand{\bx}{\bm{x}}
\newcommand{\bu}{\bm{u}}
\newcommand{\bp}{\bm{p}}
\newcommand{\bc}{\bm{c}}
\newcommand{\bpsi}{\bm{\psi}}
\newcommand{\dt}{\,{\rm d}t}

\newcommand{\T}{{^{\rm T}}}

\newcommand{\pd}[2]{\dfrac{\partial{#1}}{\partial{#2}}}

\newcommand{\sn}[2]{${#1}\times10^{#2}$}  
\newcommand{\mg}[1]{\multicolumn{2}{l}{#1}}
\newcommand{\hfour}[1]{\makecell[cl]{#1 \\ \\ \\}}
\graphicspath{{Figures/}}

\newcommand{\KHComment}[1]{}
\newcommand{\GT}[1]{\textcolor{green}{GT: #1}}

\begin{document}

\title{Discontinuity-Sensitive Optimal Control Learning by Mixture of Experts}


\author{
\authorblockN{Gao Tang}
\authorblockA{Department of Mechanical Engineering and Material Science\\
Duke University\\
Durham, North Carolina 27708\\
Email: gao.tang@duke.edu}
\and
\authorblockN{Kris Hauser}
\authorblockA{Department of Electrical and Computer Engineering\\
Duke University\\
Durham, North Carolina 27708\\
Email: kris.hauser@duke.edu}
}


%

\maketitle

\begin{abstract}
    This paper proposes a discontinuity-sensitive approach to learn the solutions of parametric optimal control problems with high accuracy.
    Many tasks, ranging from model predictive control to reinforcement learning, may be solved by learning optimal solutions as a function of problem parameters.
    However, nonconvexity, discrete homotopy classes, and control switching cause discontinuity in the parameter-solution mapping, thus making learning difficult for traditional continuous function approximators.
    A mixture of experts (MoE) model composed of a classifier and several regressors is proposed to address such an issue.
    The optimal trajectories of different parameters are clustered such that in each cluster the trajectories are continuous function of problem parameters.
    Numerical examples on benchmark problems show that training the classifier and regressors individually outperforms joint training of MoE.  With suitably chosen clusters, this approach not only achieves lower prediction error with less training data and fewer model parameters, but also leads to dramatic improvements in the reliability of trajectory tracking  compared to traditional universal function approximation models (e.g., neural networks).
\end{abstract}

\IEEEpeerreviewmaketitle

\section{Introduction}
Nonlinear Optimal Control Problems (OCPs) are critical to solve to obtain high performance in many engineering applications. For example, model predictive control (MPC) requires an OCP being solved in every control loop \cite{Bemporad2000}, while kinodynamic motion planners rely on solving OCPs between sampled states \cite{donald1993kinodynamic}.
However, they are generally difficult to solve to global optimum quickly and with high confidence due to inherent nonconvexity. 
This has led to an intense interest in using learning to obtain approximations of optimal control policies, either using supervised learning~\cite{Jetchev:2013cr,Lampariello} or reinforcement learning \cite{lillicrap2015continuous}. 

\begin{figure}[tbp]
  \centering
\begin{subfigure}[b]{0.475\columnwidth}
    \centering
    \includegraphics[width=\textwidth]{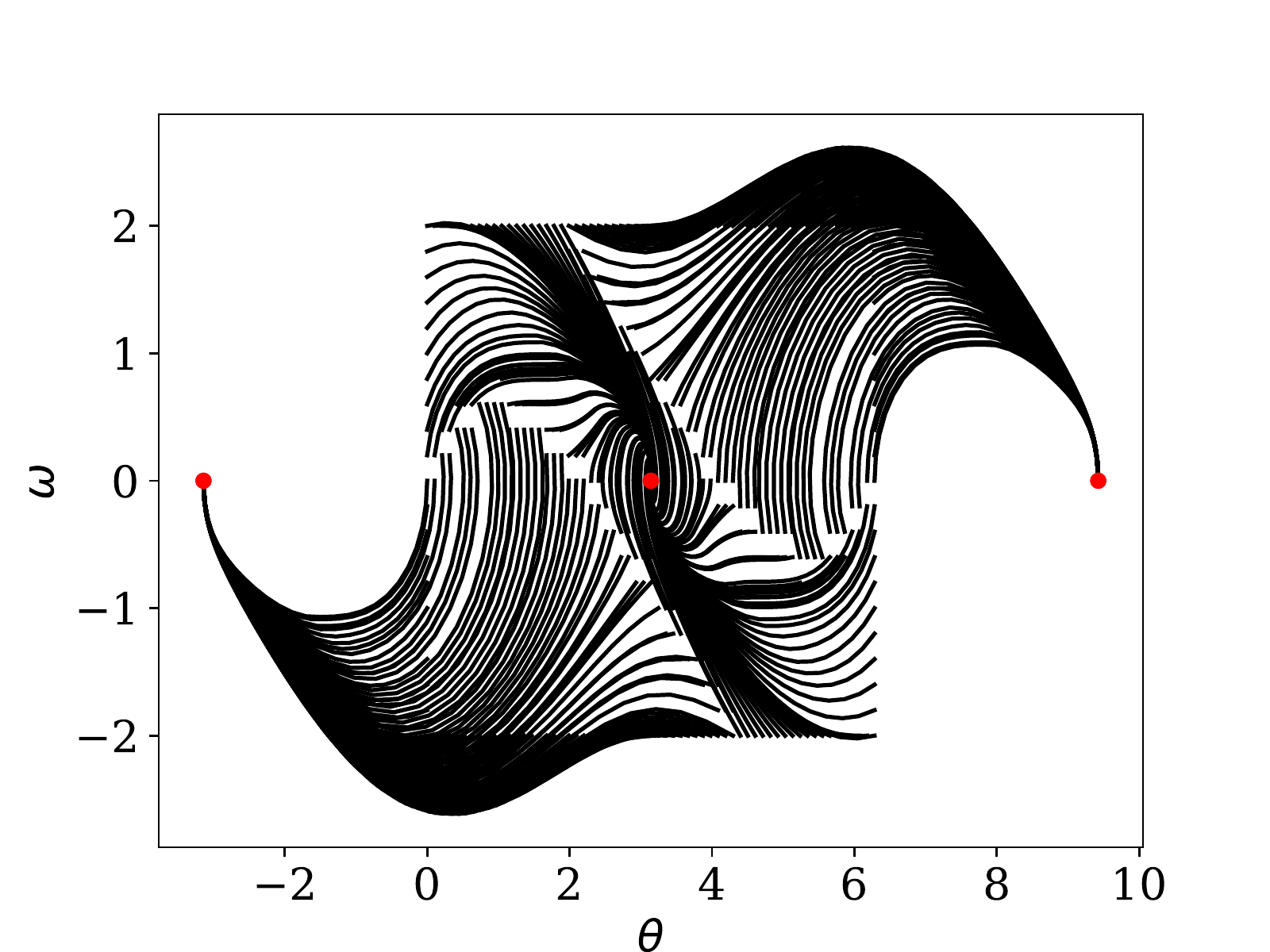}
    \caption[black pendulum]%
    {{\small Samples of data}}    
\end{subfigure}
\hfill
\begin{subfigure}[b]{0.475\columnwidth}
    \centering 
    \includegraphics[width=\textwidth]{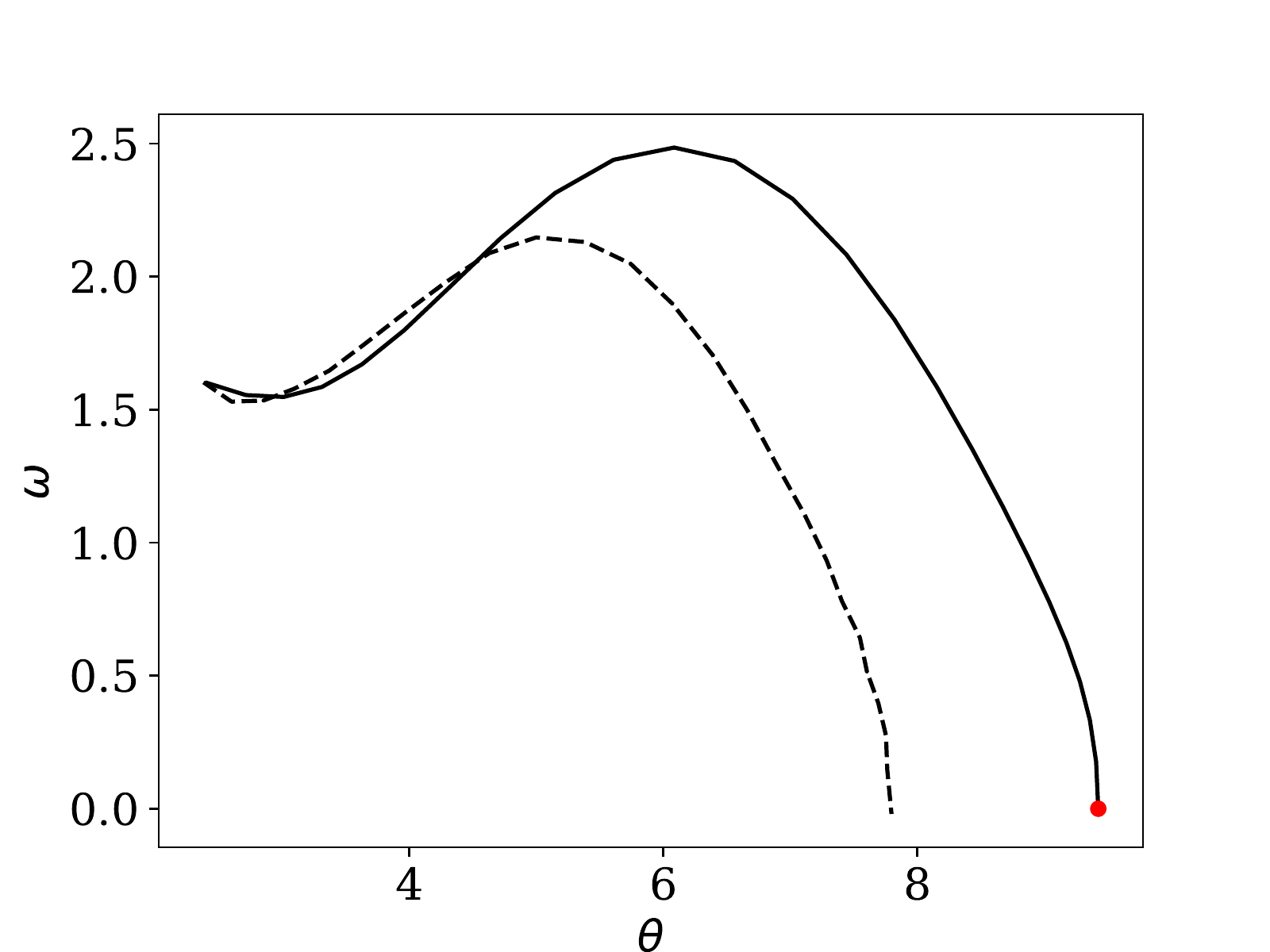}
    \caption[]%
    {{\small SNN Prediction}}    
\end{subfigure}
\vskip\baselineskip
\begin{subfigure}[b]{0.475\columnwidth}   
    \centering 
    \includegraphics[width=\textwidth]{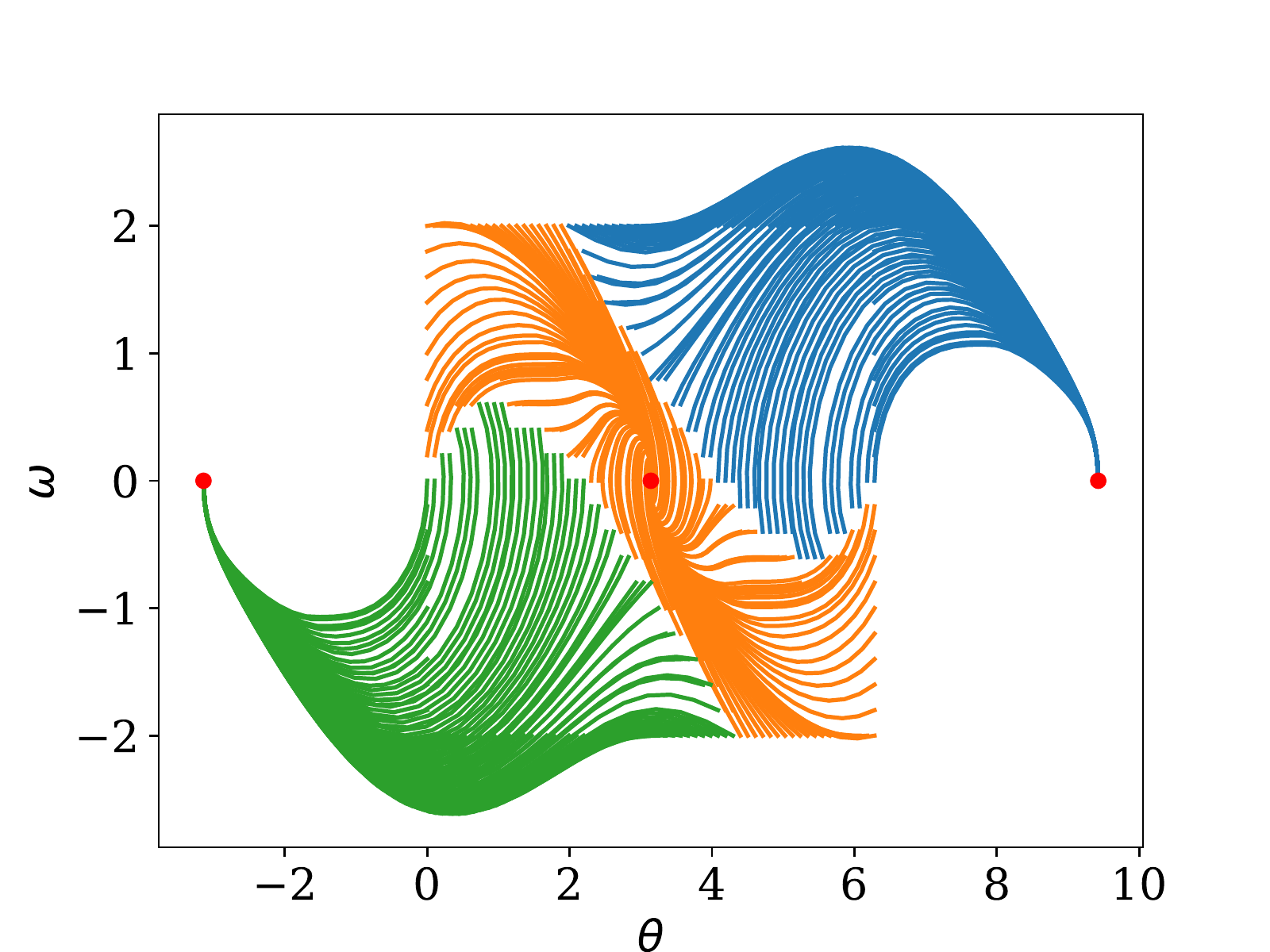}
    \caption[]%
    {{\small Samples of clustered data}}    
\end{subfigure}
\hfill
\begin{subfigure}[b]{0.475\columnwidth}   
    \centering 
    \includegraphics[width=\textwidth]{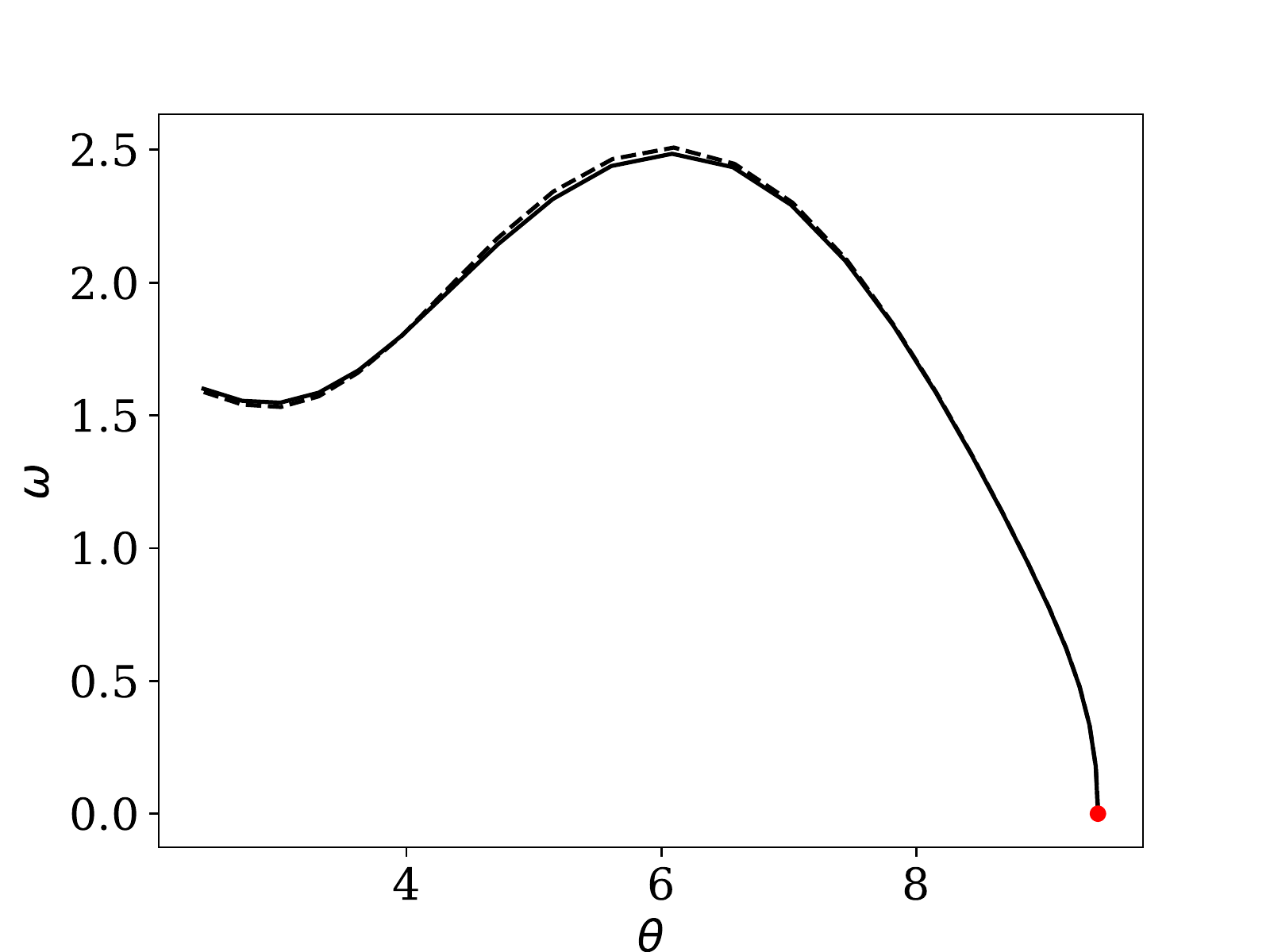}
    \caption[]%
    {{\small MoE Prediction}}    
\end{subfigure}
\caption
{\small Illustration of dataset and prediction of a selected state from SNN and MoE for the pendulum swingup task. 
(a) samples of optimal pendulum swingup trajectories from different initial states. The red circles are possible target states. 
(b) prediction of a selected state by SNN that is trained using data in (a). The solid and dashed lines denote the optimal and predicted trajectories, respectively.
(c) samples of clustered optimal trajectories where each color denotes one cluster. 
Trajectories are clustered according to final state.
(d) prediction by MoE to the same state as (b).
}
  \label{fig:SNNbadPred}
\end{figure}

In this paper, we highlight the problem that function approximators such as standard neural networks (SNN) perform poorly near discontinuities that are prevalent in many nonlinear OCPs. 
Fig.~\ref{fig:SNNbadPred} shows the results of using a multilayer SNN to learn a pendulum swingup task from optimal trajectories.
The optimal trajectories have three possible goal states so the parameter-solution mapping is discontinuous.
Although neural networks are quite useful for approximating nonlinear functions \cite{hornik1989}, near the region where the optimal goal state switches, their prediction tends to predict a final state that interpolates between two goal states.

\begin{figure}[htbp]
  \centering
  \includegraphics[width=0.5\columnwidth]{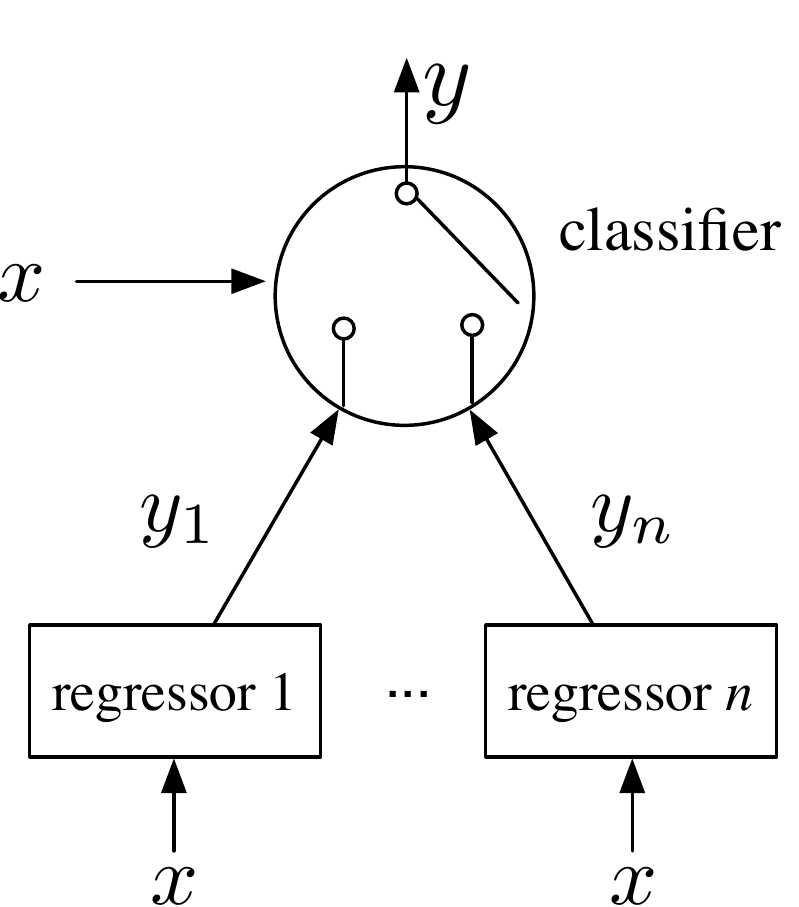}
  \caption{\small Illustration of MoE. The classifier selects a model which makes the final prediction.}
  \label{fig:showMoE}
\end{figure}

This paper addresses this problem by modifying the  Mixture of Experts (MoE) \cite{jacobs1991adaptive, jordan1994hierarchical, shazeer2017outrageously} model to learn the solutions to parametric OCPs.
The model structure uses a classifier (gating network) to select a regressor (expert) which makes the final prediction (Fig.~\ref{fig:showMoE}).
We intend to train a model such that each regressor works in a region of the parameter space where the parameter-solution mapping is continuous.  This is reminiscent of a divide and conquer approach, which has already been widely used in control community for controller design \cite{murray1997multiple}.
Fig.~\ref{fig:SNNbadPred} illustrates that the pendulum swingup dataset can be divided into three regions, and by classifying them and approximating them separately, MoE makes better prediction than SNN particularly near discontinuity.

Considerable care must be taken during MoE training.  Although MoE is generally trained using backpropagation \cite{shazeer2017outrageously} or expectation maximization \cite{jordan1994hierarchical}, training can be unstable.
We propose an approach specially designed for parametric OCPs.
The training set consists of solutions to a sampling of parametric OCPs, and we first partition the data into several clusters.
Then the classifier is trained to predict the identity of the partition and a separate regressor is trained for each partition. Each component is trained individually using backpropagation. 
Interestingly, although joint training leads to a model with lower prediction error (loss), it tends to {\em worsen} trajectory tracking success rate.
Moreover, clustering the dataset appropriately is nontrivial and it is fundamental to our approach.
Rather than using general methods of input partitioning~\cite{tang2002input}, we propose certain features of optimal trajectories that tend to work well empirically. 

Experiments on toy underactuated control problems and agile vehicle control problems demonstrate that suitably trained MoE models can learn near-optimal trajectories suitable for trajectory tracking with remarkably high success rates (99.5+\%).


\section{Related Work}
Nonconvex OCP is generally difficult to solve to global optimum, despite much work to enlarge the convergence domain, e.g., \cite{Jiang2012}.  Moreover, numerical trajectory optimization~\cite{betts1998survey} techniques are, in general, too computationally expensive for highly reactive motions.

As a result, machine learning approaches have been proposed to solve OCPs approximately but in real-time. Reinforcement learning learns the optimal policy by interacting with the environment, and deep neural network policy approximators have been shown to solve complex control problems \cite{lillicrap2015continuous}.  Another approach uses supervised learning to learn from precomputed optimal solutions to solve novel problems, and has seen successful application in trajectory optimization \cite{Jetchev:2013cr, tangdata, tomic2014} and global nonlinear optimization~\cite{Hauser2017}. 
In \cite{Jetchev:2013cr} precomputed optimal motions are used in a regression to predict trajectories for novel situations to speed up subsequent optimization.
In \cite{tangdata} the nearest-neighbor optimal control (NNOC) method is proposed, with a multiple restart method proposed to handle discontinuities. In both these works, the techniques work faster than optimizing from scratch, but still require some amount of optimization for their predicted trajectories.  This paper also learns optimal trajectories instead of optimal policies, which has the advantage that trajectories can be tracked using a stabilizing feedback controller to handle model uncertainties and disturbances.  It should be noted that the predicted trajectory might not fully satisfy the system dynamics constraints. However, if learning is sufficiently accurate, then this should not be an issue because a feedback controller can correct for such violations.

The discontinuity of the solutions to parametric OCPs as a function of problem parameters has long been known~\cite{fiacco1983}, a fact that has been underappreciated in the control learning community.
Under certain assumptions, this function is piecewise continuous, and discontinuity-tolerant methods have been proposed for learning from optimal solutions \cite{Hauser2017,tangdata}.  However, their approaches do not explicitly try to partition the space into regions. In contrast, the discontinuity-sensitive approach proposed here does indeed segment the dataset according to estimated discontinuities. 

The most related work is previous research on MoE \cite{jordan1994hierarchical, shazeer2017outrageously,tang2002input}.
This paper proposes several modifications to MoE make it suitable for learning optimal control.
We use hard classification boundaries to avoid predicting an average of both sides, and we also modify the training approach.
Traditionally MoE is trained using either backpropagation \cite{shazeer2017outrageously} or expectation maximization \cite{jordan1994hierarchical} so the gating function and experts are both updated.
However, we train the classifier and regressors individually, and experiments suggest that this is fundamental to achieving high trajectory tracking accuracy.

\section{Problem Formulation}
In this section, the problem of learning from optimal control is formulated and the key components are analyzed.
The proposed approach first forumlates a parametric OCP and then performs the following procedure:
\begin{enumerate}
    \item Input: collect dataset of solutions to parametric OCPs on sampled parameters.
    \item Cluster: select a clustering approach to cluster the trajectories and partition the parameter space.
    \item Train: weights of classifier and regressors are trained individually using backpropagation.
    \item Validate: predict optimal trajectories for novel states and validate the learned model by trajectory rollout.
\end{enumerate}
\subsection{Parametric Optimal Control}
A system is governed by dynamical equations 
\begin{equation}
    \label{eq:contsysdyn}
    \dot{\bs{x}}=\bs{f}(t,\bs{x}, \bs{u}, \bs{p})
\end{equation}
where $t$ is time; $\bs{x} \in \mathbb{R}^n$ is the state variable; $\bs{u} \in \mathbb{R}^m$ is the control variable; $\bs{p}\in \mathbb{R}^l$ is the problem parameters and captures the variability of studied problems. The vector $\bs{p}$ may specify the initial state, model parameters, and modifications to costs or constraints.
We use subscript 0 and $f$ to denote the variables at initial and final time, respectively. 
The goal is to control the system from some state $\bx _0$ to some state $\bx _f$ while minimizing the cost function
\begin{equation}
    \label{eq:contJ}
    J=\varphi(t_0, \bx _0, t_f, \bx _f, \bp) + \int_{t_0}^{t_f}L(t, \bx(t), \bu(t), \bp)
\end{equation}
where $\varphi$ only depends on initial and final states; $L$ depends on state and control variables within $[t_0, t_f]$.
Practical OCPs may have state, control, and terminal set constraints that have to be satisfied and we refer to \cite{betts1998survey} for details. 

Parametric OCP is generally difficult to solve analytically~\cite{Maurer:2001by},
but for any given parameter, numerical methods may be used to solve the resulting OCP~\cite{betts1998survey}.
In this work we employ a direct transcription method, which transforms the OCP into a nonlinear optimization problem  and solves it using SNOPT~\cite{gill2005snopt}.
The solution trajectory is a sequence of state and control variables along a time grid, denoted as $\bm{z}\equiv\{t_i;\bx_i;\bu_i\}_{i=0}^{N}$ where $N$ is the grid size for discretization.  Stacking the element of $\bm{z}$ into a vector, our goal is to approximate the map from problem parameters to optimal trajectories $\bm{z}^\star(\bs{p})$.

\subsection{Optimal Trajectory Database Generation}

To train and test models we generate a database of optimal trajectories $\bm{z}_1,\ldots,\bm{z}_M$ to sampled problems $\bs{p}_1,\ldots,\bs{p}_M \in \mathbb{R}^l$.
Due to non-convexity, even finding a global optimum to a single problem can be difficult. One practical approach is to pick the best local optimum from a multi-start method.
However, the local optimum can be also quite difficult to find if an initial guess not close to the optimum is provided.
We adopt a nearest-neighbor approach \cite{tangdata} to help generate large databases quickly.  We first sample some number of problems (fewer than $M$ but much larger than the number of expected partitions) and use an exhaustive random restart approach to solve them.  These solutions are used as the initial database.  Then we sample more parameters, and for each new problem we attempt local optimization from each of its $k$-nearest neighbors to find $k$ local optima.  The best solution is kept in the database.
We note that this process is done completely offline and parallelizable.

\subsection{Mixture of Experts}

The MoE model is composed of a classifier and $r$ regressors, as shown in Fig.~\ref{fig:showMoE}.  In this paper both models are chosen as multilayer perceptrons (MLP).
The goal is to learn a function $z:\mathbb{R}^l \to \mathbb{R}^R$ that approximates $\bm{z}(\bs{p})$ where $R$ is the length of vector $\bm{z}$.
Each regressor takes input $\bs{p} \in \mathbb{R}^l$ and makes a prediction $y_i(\bs{p}, w_i) \in \mathbb{R}^R, i=1,\dots,r$ where $w_i$ specifies the weights of each regressor. 
The classifier, with weights $w_c$, takes input $\bs{p}$ and predicts $r$ values $\{c_i\}_{i=1}^r$.
The output of the classifier are combined with softmax to assign probabilities for each model, i.e.
\begin{equation}
    P_i=\frac{\exp{c_i}}{\Sigma_{i=1}^{n}\exp{c_i}}
\end{equation}
or argmax to select one model only (in this case, $P_k=1$ for $k=\arg\max_i c_i$ and $P_k=0$ otherwise.)  The difference between softmax and argmax is softmax tends to give a prediction that is a mixture of predictions from all experts. 
Argmax, however, selects one model and ignores other models' predictions.

In either case, the ultimate prediction is a mixture of predictions from all regressors, i.e. 
\begin{equation}
z(\bs{p})=\Sigma_{i=1}^n P_i(\bs{p},w_c) y_i(\bs{p}, w_i) 
\end{equation}
The target is to find $w_c$ and $\{w_i\}_{i=1}^r$ in order to miminize
\begin{equation}
    \label{eq:trainobj}
    L=\mathbb{E}_{\bs{p}\sim P_{\text{data}}}\text{loss}(z(\bs{p}), \bm{z}^\star(\bs{p}))
\end{equation}
where $P_{\text{data}}$ is a distribution over problems and $\text{loss}(\cdot, \cdot)$ is any regression loss function.

The most straightforward way train MoE is to treat it as an SNN, randomly initialize weights, and miminize \eqref{eq:trainobj} using backpropagation. Although several heuristics have been proposed to train MoE using backpropagation such as \cite{shazeer2017outrageously}, training may still be unstable.
If softmax is used, all the data is used to train each regressor, with weights equal to the probabilities predicted by the classifier.  In the case of argmax, each regressor is only trained using data assigned to it by the classifier.
There is no gradient for the classifier to update its weights if argmax is used.
Softmax, on the other hand, can still have gradient to update the weights of the classifier. 

To perform joint training, since argmax is the limit of softmax if we scale $\{c_i\}_{i=1}^{r}$ by a large positive scalar, we introduce $\epsilon\in[0, \infty)$ which is used to divide the output of the classifier before applying softmax, i.e.
\begin{equation}
    P_i=\frac{\exp{(c_i/\epsilon)}}{\Sigma_{i=1}^{n}\exp{(c_i/\epsilon)}}.
\end{equation}
As $\epsilon\rightarrow 0$, the softmax weights approach the argmax function.  Hence, $\epsilon$ must be gradually lowered to balance between updating weights of classifier and restricting mixture of outputs from multiple regressors.  As we shall show later, joint training of MoE may improve the loss function compared to decoupled training, but appears to be detrimental to trajectory tracking performance. 


\subsection{Parameter Space Partition}
Clustering has been shown to be effective to avoid some instability in MoE training~\cite{tang2002input} by training the classifier and regressors of MoE individually on subsets of the data.  We adopt the same approach here, and study how to partition parameter space such that in each region the parameter-solution mapping is continuous.

The dataset $\{(\bs{p}_j, \bm{z}_j)\}_{j=1}^M$ is divided into $r$ groups $C_1,\ldots,C_r$, ideally so that $\bm{z}^\star(\bs{p})$ is a continuous function for all $\bs{p}$ in a given region.  
This problem can be formulated as a clustering problem and each cluster denotes a region of the partitioned parameter space.  The classifier is trained to predict $P_i(\bs{p}_j,w_c)=1$ for all $\bs{p}_j$ in $C_i$, and the $i$'th regressor is trained as usual, restricted to the examples in $C_i$.  We call this process (decoupled) {\em pretraining}.

Parametric OCPs have rich features that can be used to find appropriate clusters.
We note that this partition cannot be done simply using problem parameters only since the target is to find the discontinuity in the solutions.
Discontinuity comes from switching from one family of local optima to another.  Hence, although the objective function value and the problem parameters at these discontinuities is similar, the trajectory may not.  For example, a car might reverse first or move forward first, and a quadcopter might avoid an obstacle from above or below.

Hence, we experiment with using distance between optimal trajectories to classify the family of solution.  The simplest approach is to apply standard clustering techniques, such as the k-Means algorithm, on the trajectory vector space.  In order to do so, we first normalize the state and control variables to zero mean and unit variance.  After choosing a number of clusters $k$, the k-Means algorithm is run from random initial centers.

Our experiments observe that k-Means is for some problems successful at predicting discontinuities, but can also group trajectories poorly when $k$ is small.  On the other hand, when $k$ is large, each cluster contains less training data, causing the regressors to overfit, and making the job of the classifier harder.  

We also propose custom clustering criteria that are based on a system designer's intuition and inspection of datasets.  As an example, the periodicity of angles is a useful feature when angle is in the state space and optimal trajectories have distinct final angles; in other words, trajectories lie in distinct homotopy classes.
This is useful for the pendulum swingup problem as well as the ground vehicle control problem we consider later.  Another approach that is applicable is to examine the Lagrange multipliers of constraints at optimal solutions, since they provide rich information about how constraints influence the trajectory's shape.
For example, in quadcopter obstacle avoidance the shortest path might go on either side of the obstacle.  Hence, the gradient of the active constraint will have different sign.

\subsection{Discussion and Preliminary Experimentation}

The usual approach to MoE is to first perform pretraining before (coupled) {\em retraining} by minimizing~\eqref{eq:trainobj}.
The rationale is that pretraining provides a good initialization, but if the data is clustered badly, i.e. in one cluster there is discontinuity, the loss function may be large.  Moreover, even if clustering is perfect, a pretrained model does not necessarily minimize \eqref{eq:trainobj} due to misclassification.  In this section we shall experimentally demonstrate and discuss why this may be a poor approach for parametric OCPs. 

\label{sec:RolloutError}

\begin{figure*}[tbp]
  \centering
\begin{subfigure}[b]{\columnwidth}
    \centering
    \includegraphics[width=\textwidth]{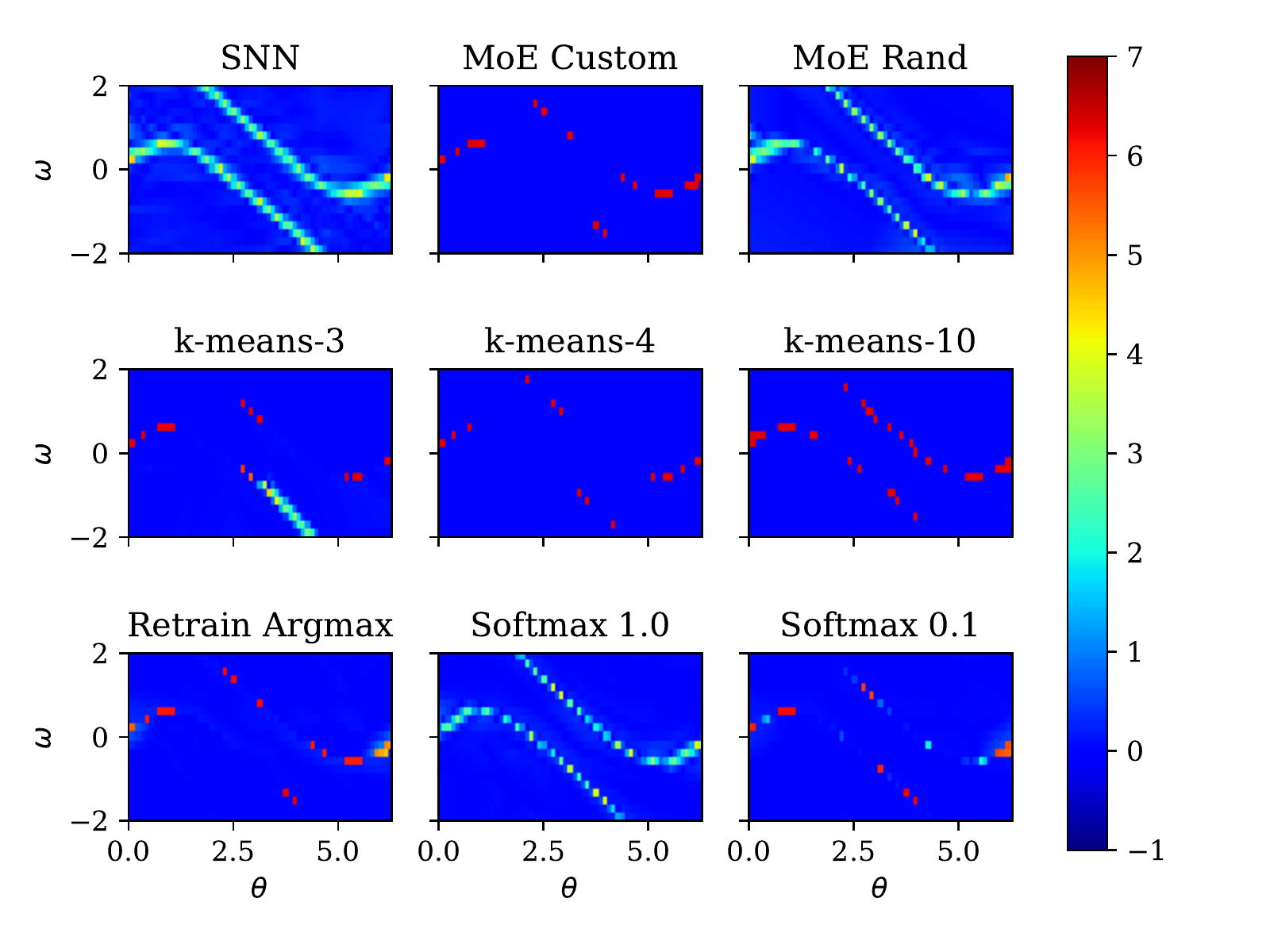}
    \caption
    {{\small Prediction error of $\theta_f$}}    
\end{subfigure}
\hfill
\begin{subfigure}[b]{\columnwidth}  
    \centering 
    \includegraphics[width=\textwidth]{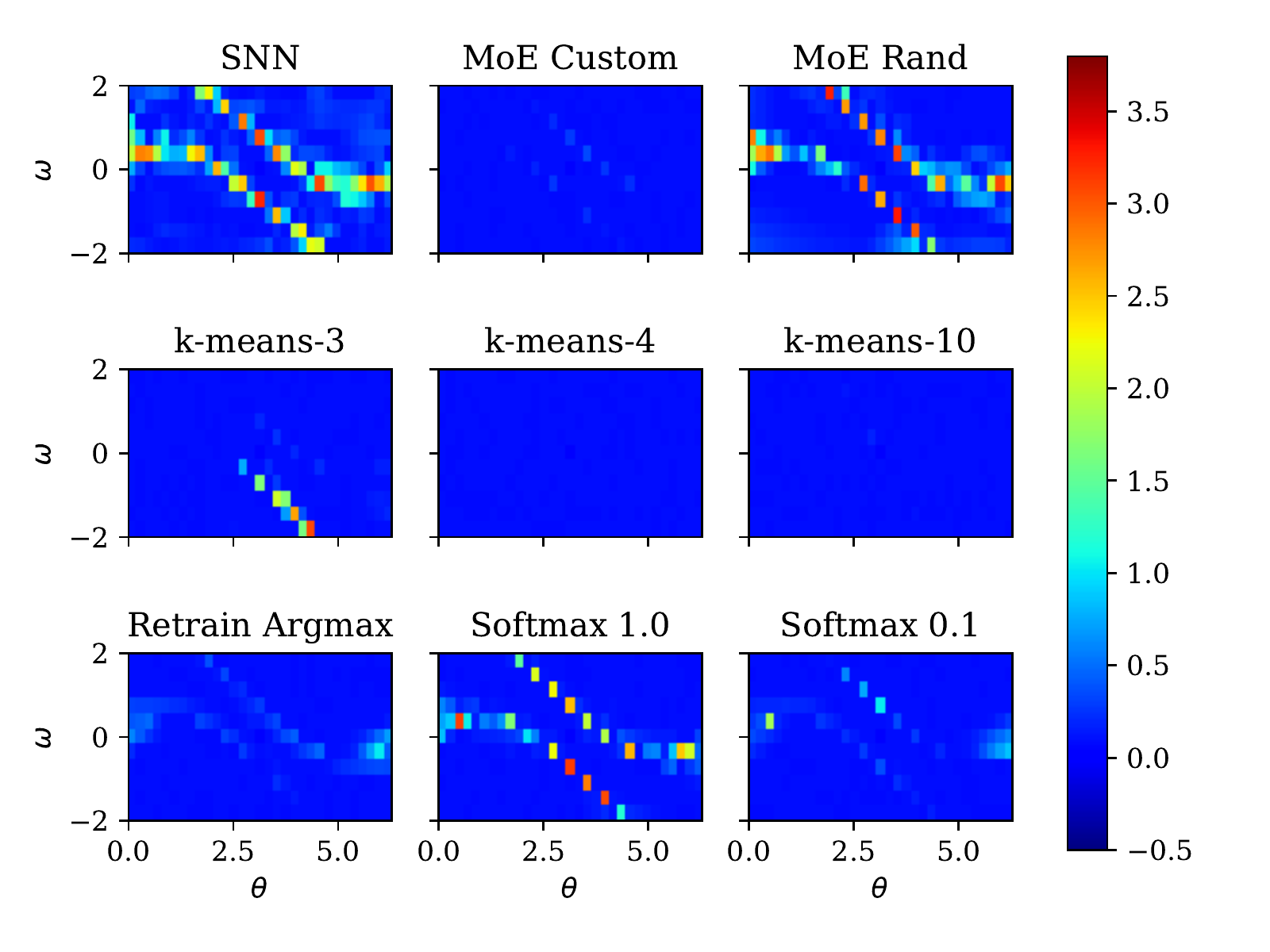}
    \caption
    {{\small State error after trajectory tracking}}    
\end{subfigure}
\caption{\small Comparing several models for learning the pendulum swing-up task. 
}
  \label{fig:penLargeError}
\end{figure*}

We study a toy pendulum swingup task, where the task is to reach the upright position.  Details on the system and neural network models are given in Sec.~\ref{sec:Pendulum}.  
We compare on two metrics: 1) test error (smoothed L1 loss) and 2) rollout success rate after trajectory tracking.  In trajectory tracking, we simulate trajectory execution under an LQR controller, which compensates for errors dynamic constraint violations.  About each state along the predicted trajectory, we compute an LQR solution for a linear dynamics model and a quadratic cost obtained by Taylor expansion.  After trajectory tracking is complete, the simulation switches to a stabilizing controller about the origin. If after 5 seconds the norm of the state error is within a certain threshold (0.1) we denote the rollout as a success. (We note that for the car problem, only the first stage is implemented since the final state is not controllable.)

The following variations are considered:
\begin{enumerate}
    \item SNN vs MoE,
    \item MoE with random weights against $k$-means clustering on trajectories, and against custom clustering, and
    \item Retraining vs no retraining.
\end{enumerate}
The SNN is chosen as MLP of size (2, 300, 75), there the first number denotes the size of the input layer, the last number denotes the size of the output layer, and intermediate numbers indicate the size of hidden layers.  We experimented with SNN with more hidden layers or more neurons in the hidden layer, but they result in similar or larger test error.
Specifically, MLPs of size (2, 50, 20, 75) and (2, 20, 50, 75), (2, 30, 30, 30, 75) yield test errors of 0.258, 0.170, and 0.232, respectively. The size (2, 300, 75) network, on the other hand, has test error of 0.058.

For MoE, the classifier is of size (2, 50, $r$) and the $r$ regressors are all of size (2, 20, 75).  Custom MoE and random weight MoE use $3$ experts. The custom clustering divides the data into 3 clusters based on the final angle.  
We also use $k$-means with 3, 4, and 10 clusters solely on trajectories with the same design of network size.



Fig.~\ref{fig:penLargeError}.a plots the prediction error on $\theta_f$ and Fig.~\ref{fig:penLargeError}.b plots the state error after trajectory tracking.
The validation error and rollout success for each model are also listed in Tab.~\ref{tab:cmpTraining}.  

Row 1 shows that SNN has difficulty in making predictions in regions near the discontinuity, averaging between both sides.
MoE does also make inaccurate prediction, but these are caused by misclassification and the prediction is a local optimal trajectory belonging to another cluster.  Hence, they are suboptimal but still reach the vertical position as desired, since the difference in $\theta_f$ is $2\pi$.
The suboptimality is not too great, because near the boundaries two families of solutions have similar objective function.
MoE trained from random initialization does achive lower prediction error than SNN, but is not very successful.
 This indicates that training by simply descending \eqref{eq:trainobj} is unable to guide the classifier to the appropriate clusters.

\begin{figure}[tbp]
  \centering
  \includegraphics[width=\columnwidth]{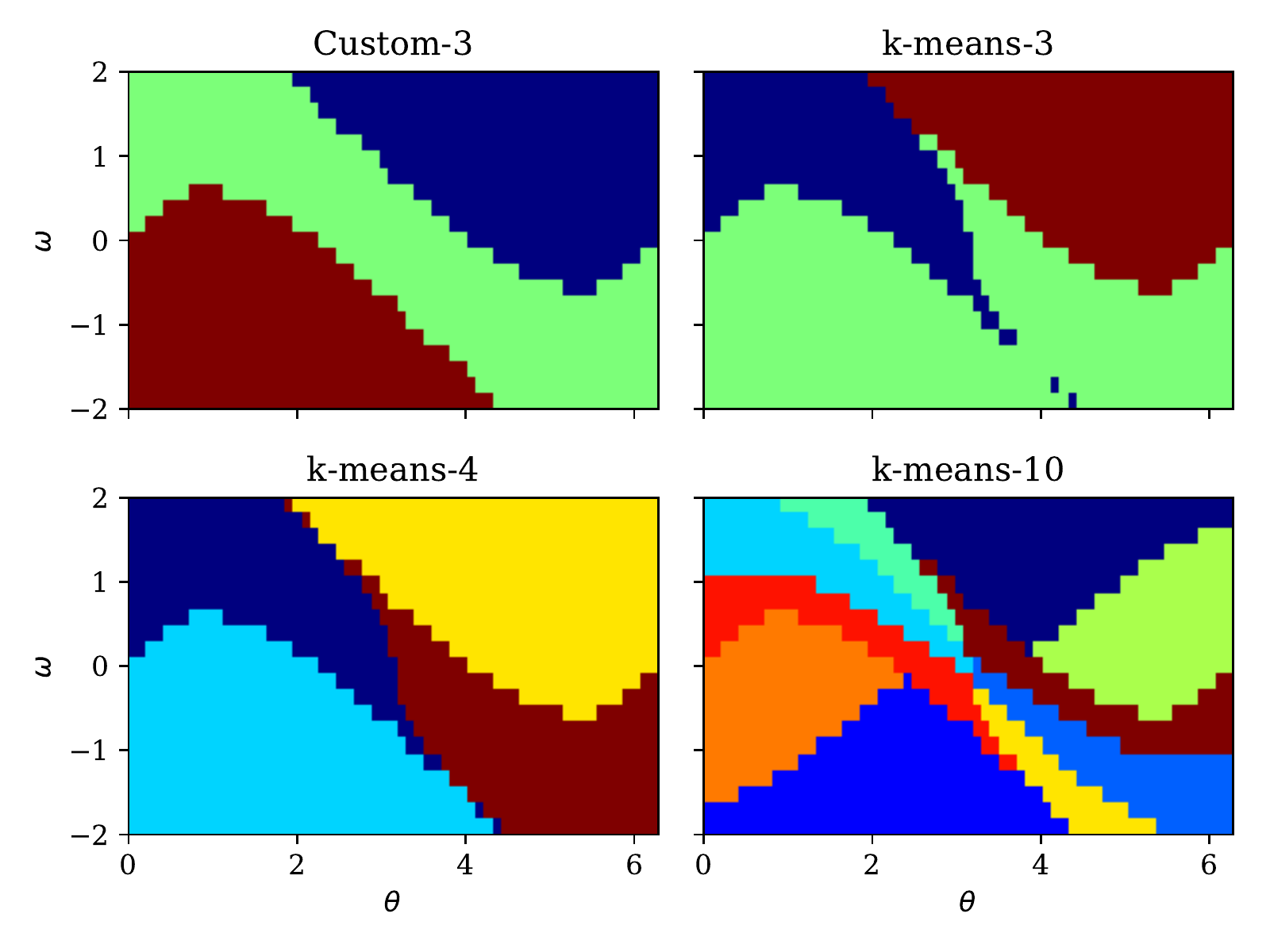}
  \caption{\small Choices of clusters for the pendulum problem. Different colors means different clusters. Figures include: 1) custom 3 clusters 2) k-Means with 3 clusters 3) k-Means with 4 clusters 4) k-Means with 10 clusters}
  \label{fig:penClusters}
  \vspace{-0.5cm}
\end{figure}

Row 2 tests MoE with k-Means and various cluster sizes., which are shown in Fig.~\ref{fig:penClusters}. 
$k=3$ has one cluster that has data from both families of trajectories, so the prediction close to the discontinuity is worse. $k=4$ and $k=10$ clusters finds the discontinuity successfully, and the resulting MoE achieves high success rate.

Row 3 of Fig.~\ref{fig:penLargeError} shows various methods of retraining after pretraining MoE with custom clustering. In all cases this approach decreases regression error but also rollout success rate. 
In (vii) argmax is used following the output layer of the classifier. 
The classifier has no gradient to update it self so only the regressors are updated.
Due to classification error, the regressors will be trained with trajectories from other clusters.
As a result, the prediction near the boundaries will tends towards the average of two clutsers.
In (vii) and (ix) we use softmax with different $\epsilon$.
In these cases, the classifier is updated but the regressors will predict towards the average.
As shown in Tab.~\ref{tab:cmpTraining}, retraining does decrease the prediction error at the cost of lower rollout success rate.

\begin{table*}[tbp]
\caption{\small Comparison of prediction error and rollout success rate on the pendulum problem}
\label{tab:cmpTraining}
\begin{center}
\begin{tabular}{@{}llllllllll@{}}\toprule
Model  & SNN & \multicolumn{8}{c}{MoE} \\
Clustering & --- & Custom  & Rand. & k-means-3 & k-means-4 & k-means-10 & Custom & Custom & Custom \\
Retrain & --- & --- & --- & --- & --- & --- & argmax & softmax 1.0 & softmax 0.1 \\ \midrule
    Validation error &  0.046& 0.030& 0.035& 0.039& 0.029& 0.051& 0.027& 0.028& 0.026\\ 
    Success (out of 1000) & 717& 998& 829& 970& 1000& 1000& 941& 896& 969\\ 
\bottomrule
\end{tabular}
\vspace{-0.5cm}
\end{center}
\end{table*}

These experiments suggest that proper clustering is important for MoE training.  Moreover, rollout success is a better metric to use in practice, while testing error can be misleading.  Due to misclassifications, a lower testing error can be achieved by averaging at discontinuities, but this leads to severe failures.
We also observe that coupled retraining is detrimental to performance.  This is because the imperfect classification causes the individual regressors to be provided with discontinuous training data, again leading to averaging artifacts.

\section{Numerical Examples}
We run experiments on the pendulum task and three dynamic vehicle problems, and the details are given below. Results are summarized in Tab.~\ref{tab:summary}.  In each case, training sets contained 80\% of examples specified in Dataset size, and the testing sets had the remaining 20\%.  Validation sets (of size Validation size) are generated separately.

SNN test error indicates the testing error when training is terminated. SNN hyperparameters (SNN size) were tuned to achieve low test error. Validation error (SNN/MoE validation) indicates loss on the validation set, while rollout error (SNN/MoE rollout) indicates success rate during trajectory tracking.  Except for the car problem, this involves the stabilizing LQR approach described in Sec.~\ref{sec:RolloutError}. Details on the car rollout success criteria are specified below. 

Details on the MoE network design are listed in the rows listing the number of clusters, the resulting cluster sizes, and the network hyperparameters (Classifier/Regressor size). The Regressor Test Error row indicates how well the MoE regressors are fitting on clustered data, showing that each regressor has quite small error when fit on a continuous region.

In all of these experiments, hidden layers use LeakyReLU with $\alpha=0.2$.
The output layer of regressors is a linear layer without nonlinear activation function.
The loss function is the smooth L1 loss and cross entropy loss for regressors and classifier, respectively.

\subsection{Pendulum Swing-up}
\label{sec:Pendulum}
\subsubsection{Problem Setup}
The system dynamic equations are
\begin{equation}
        \dot{\theta}=\omega,        \dot{\omega}=u-\sin\theta
\end{equation}
where $\theta, \omega$ are the angle and angular velocity of the pendulum; $u\in[-1, 1]$ is the control torque.
The problem parameters are the initial states.
The target state is the straight up state, i.e. $\omega_f=0, \mod(\theta_f, 2\pi)=\pi$.
The cost function is a weighted sum of time and control energy, i.e. $J=w(t_f-t_0)+r\int_{t_0}^{t_f}u^2 \dt$ with $w=1, r=1$.
\subsubsection{Data Generation and Training}
The parameter space is a subset of $\mathbb{R}^2$ and we directly sample parameters on a uniform grid. Specifically, we use a grid size of $61\times21$. The validation set is sampled at random.
Samples of optimal trajectories are shown in Fig.~\ref{fig:SNNbadPred}. 
The custom clustering partitions the trajectories by $\theta_f$.

\begin{table*}[htbp]
\centering
\begin{threeparttable}
\caption{\small Summary of experimental results for SNN and MoE}
\label{tab:summary}
    \begin{tabular}{@{}lllllll@{}}\toprule
        & pendulum & \multicolumn{2}{l}{ground vehicle} & quadcopter & \multicolumn{2}{l}{quadcopter-obstacle} \\ \midrule
        State dims & 2 & \mg{4} & 12 & \mg{12} \\
        Control dims & 1 & \mg{2} & 4 & \mg{4} \\
        Problem param. & $\bx_0\in\mathbb{R}^2$ & \mg{$\bx_0\in\mathbb{R}^4$} & initial position, $\mathbb{R}^3$ & \mg{\makecell[cl]{initial position and \\obstacle, $\mathbb{R}^7$}} \\
        Param range & $[-\pi, \pi] \times [-2, 2]$ & \mg{$[-10, 10]^2\times[-\pi,\pi]\times[-3.1, 3.1]$} & $[-10,10]^3$ & \mg{$[-10, 10]^6\times[1, 5]$\tnote{a}}\\
        Dataset size\tnote{\textdagger} & 1281 & \mg{120009} & 9000 & \mg{616758} \\
        Validation size & 1000 & \mg{10000} & 1000 & \mg{10000}\\
        SNN size & (2, 300, 75) & \mg{(4, 200, 200, 149)} &  (3, 200, 317) & \mg{(7, 1000, 1000, 317)}\\
        SNN test error & 0.058 & \mg{0.045} & \sn{8.6}{-5} & \mg{0.014} \\
        {\bf SNN validation} & 0.046 &  \mg{0.046} & \sn{4.7}{-5} & \mg{0.024} \\
        {\bf SNN rollout} & 717/1000 & \mg{6729/10000} & 1000/1000  & \mg{-0.315 \tnote{b}}\\
        \# clusters & 3 & \mg{6} & 4 & \mg{8}\\
        cluster approach & custom & custom & k-means & k-means & custom & kmeans\\
        Cluster size range & [388,505] & [7266,45626] & [7228, 28913]  & [2072,2356] & [70474,84280] & [64682, 101669] \\
        Classifier size & (2, 50, 3) & (4, 200, 6) & (4, 200, 6) & (3, 50, 4) & (7, 200, 200, 8) & (7, 200, 200, 8)\\
        Test accuracy & $97.2\%$ & $98.7\%$ & $97.7\%$ & $99.6\%$ & $88.9\%$ & $96.4\%$ \\
        \hfour{Regressor size} & \hfour{(2, 20, 75)} & \makecell[cl]{(4, 200, 149) \\ for small clusters\\ (4, 500, 149) \\ for large} & \makecell[cl]{(4, 200, 149)\\ for small clusters\\(4, 300, 149)\\for large} & \hfour{(3, 50, 317)} & \hfour{(7, 200, 200, 317)} & \hfour{(7, 200, 200, 317)}\\
        Regressor test error  & 0.0032 $\pm$ 0.0031 & 0.0018 $\pm$ 0.0014 & 0.0088 $\pm$ 0.0082 & \sn{4.6}{-5} $\pm$ \sn{9}{-6} & 0.0022 $\pm$ 0.0003 & 0.0052 $\pm$ 0.0027 \\
        {\bf MoE validation} & 0.030 & 0.019 & 0.031 & \sn{4.6}{-5} & 0.015 & 0.016\\
        {\bf MoE rollout} & 998/1000 & 9975/10000 & 9413/10000 & 1000/1000 & -0.043\tnote{b} & -0.167\tnote{b}\\
\bottomrule
\end{tabular}
\begin{tablenotes}
\item[a] The obstacle is sampled such that it always collides with optimal obstacle-free trajectory
\item[b] Average of the largest constraint violations based on trajectory rollout. All states can be controlled to the target.  See histogram in Fig.~\ref{fig:constrVio} for distribution.
\end{tablenotes}
\end{threeparttable}
\end{table*}

\subsection{Ground vehicle}
\subsubsection{Problem Setup}
We use a planar car with dynamic equations
\begin{equation}
\dot{x}=v\sin\theta, \,
  \dot{y}=v\cos\theta, \,
  \dot{\theta}=u_\theta v, \,
  \dot{v}=u_v
  \label{eq:CarDyn}
\end{equation}
where the state $\bs{x}=[x,y,\theta,v]$ includes the planar coordinates, orientation, and velocity of the vehicle; the control $\bs{u}=[u_\theta,u_v]$ includes the control variables which change the steering angle and velocity, respectively. 
The problem parameters are the initial states, as listed in Tab.~\ref{tab:summary} and the goal is to control the system to the origin with zero velocity and $\mod(\theta_f, 2\pi)=0$.
The cost function is a weighted sum of time and control energy, i.e. $J=w(t_f-t_0)+\int_{t_0}^{t_f}r_1u_\theta^2+r_2 u_v^2 \dt$ with $w=10, r_1=r_2=1$.
\begin{figure}
\centering
\begin{subfigure}[b]{0.475\columnwidth}
    \centering
    \includegraphics[width=\textwidth]{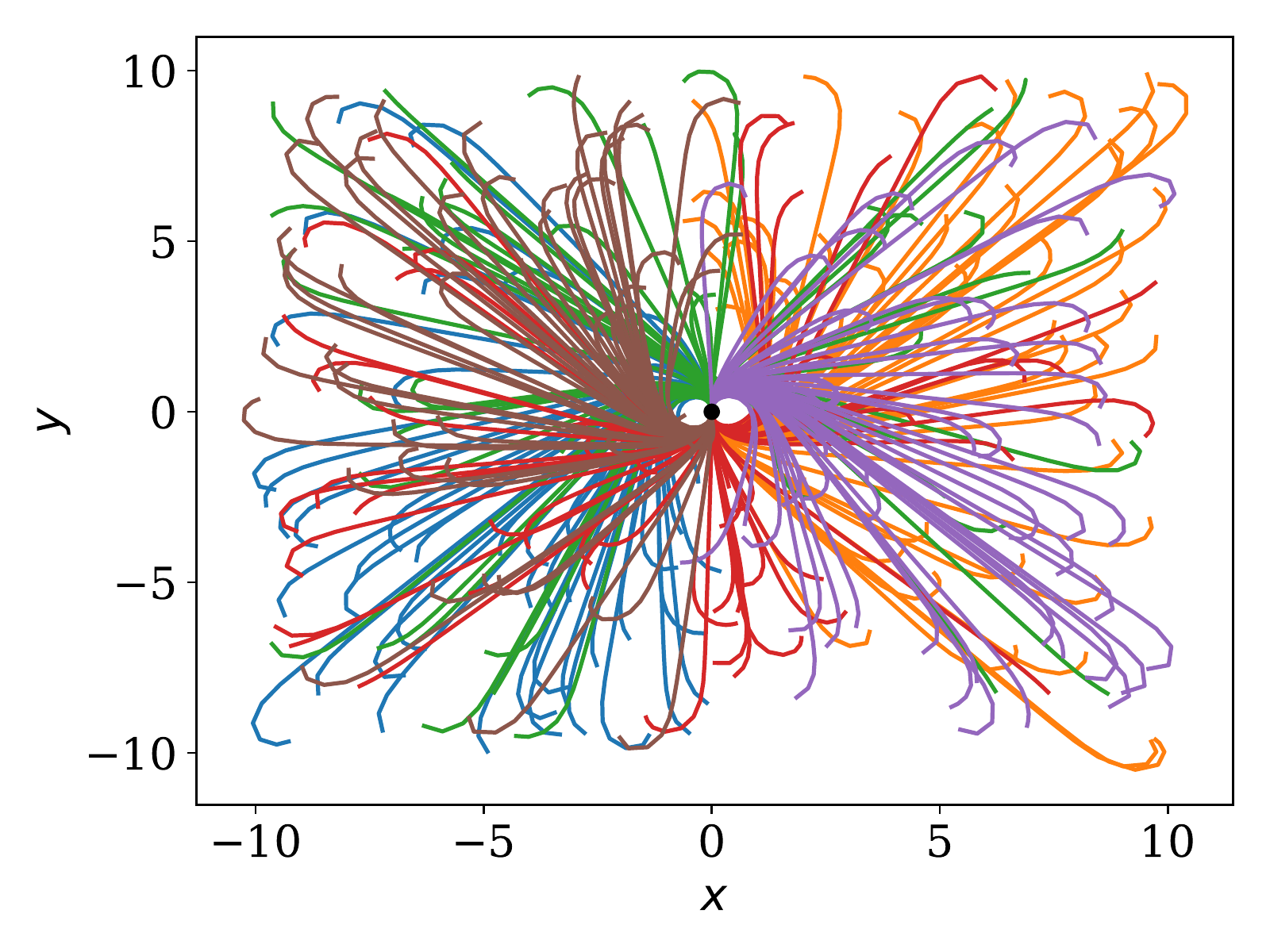}
    \caption[MoE Prediction]%
    {{\small Samples of optimal trajectories for the car problem}}    
\end{subfigure}
\hfill
\begin{subfigure}[b]{0.475\columnwidth}  
    \centering 
    \includegraphics[width=\textwidth]{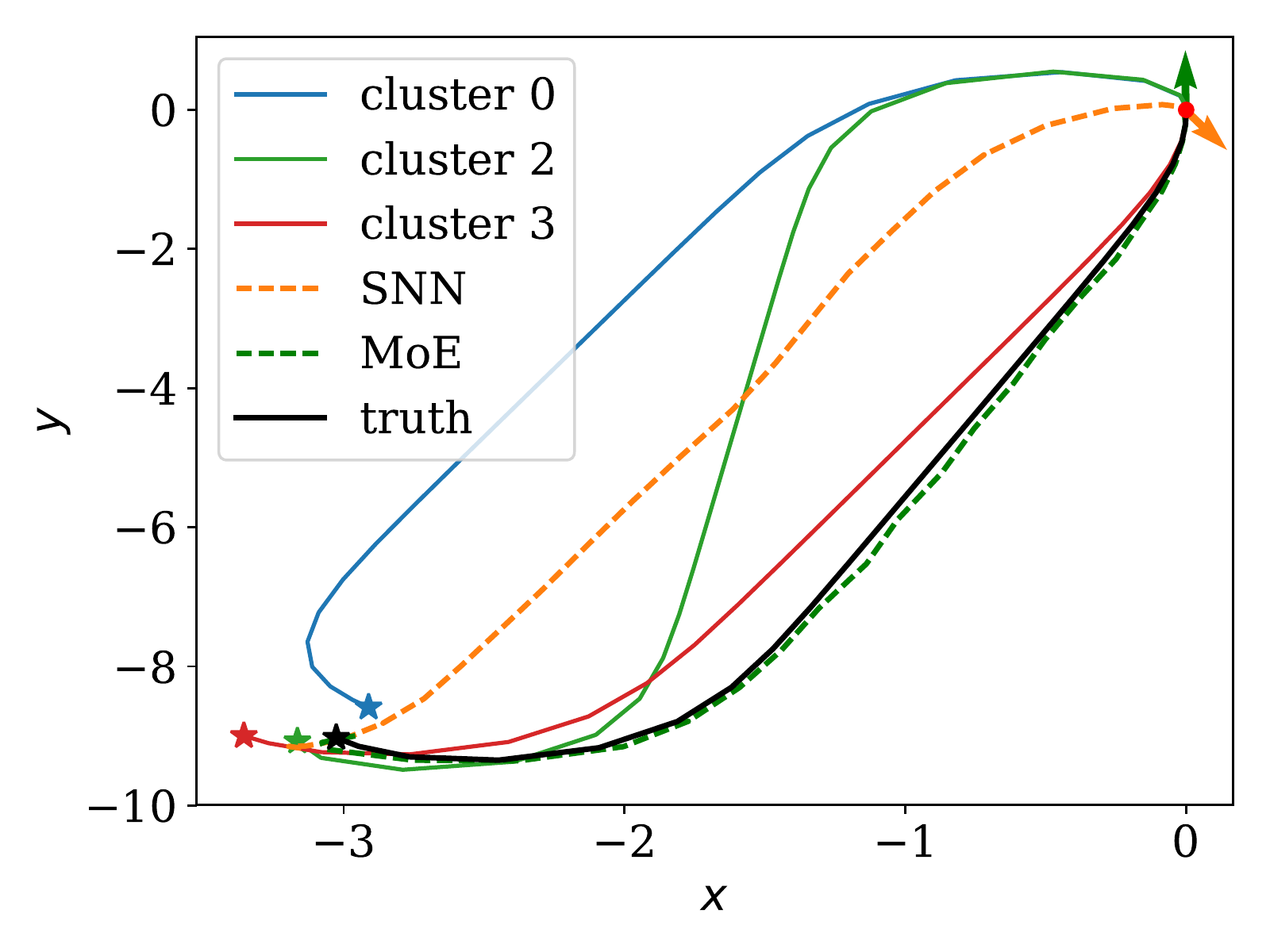}
    \caption[]%
    {{\small Illustration of poor predictions from SNN}}    
\end{subfigure}
\caption
{\small Left: samples of optimal trajectories. Each color corresponds to one cluster of trajectories. Black circle is the target. Right: A selected state that SNN makes worse prediction than MoE. It also shows states near this state might belong to to three different trajectory clusters. SNN predicts a trajectory with incorrect final angle. 
} 
\label{fig:carSNNbadPred}
\end{figure}
\subsubsection{Data Generation and Training}
\begin{figure}[tbp]
  \centering
  \includegraphics[width=\columnwidth]{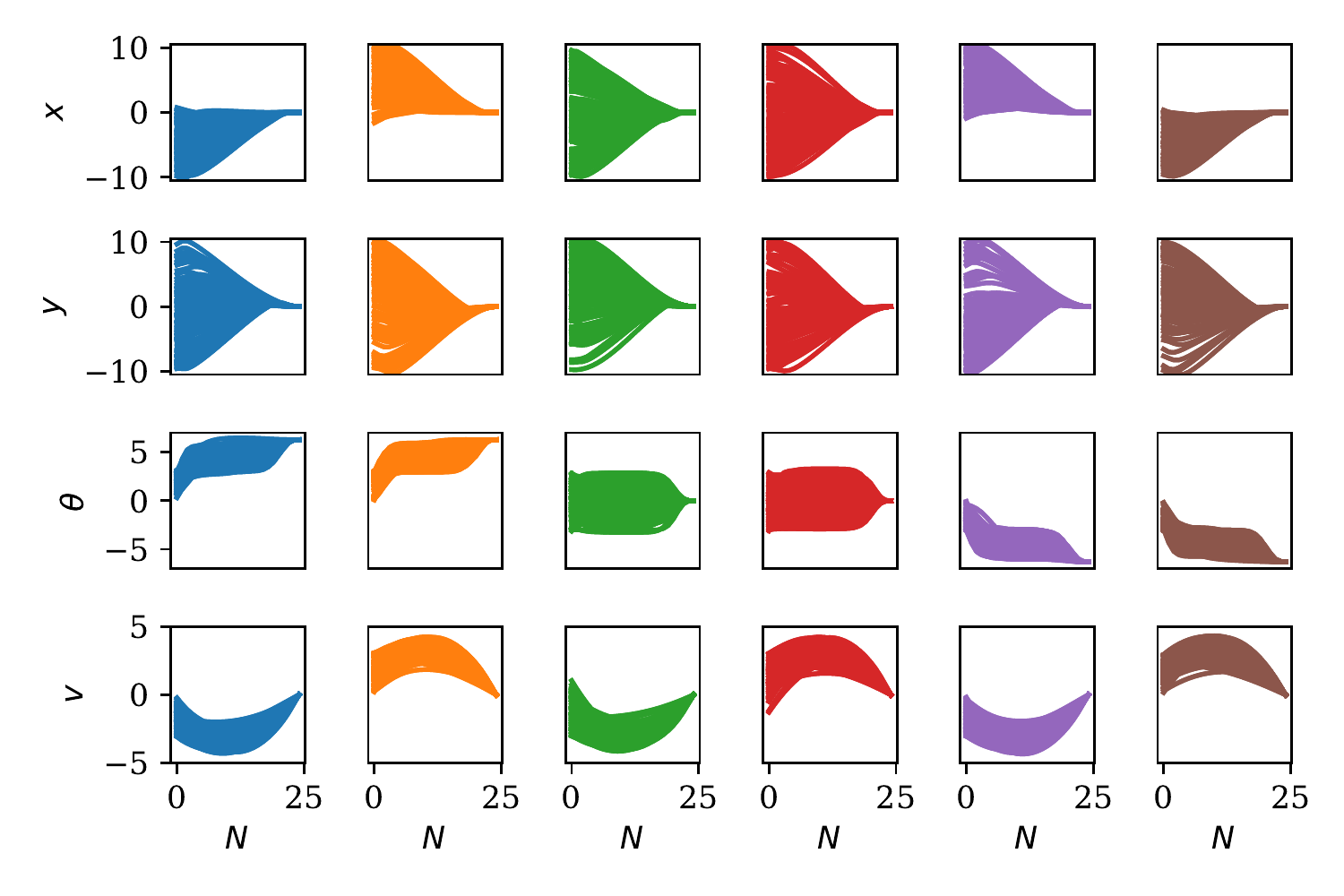}
  \caption{\small Samples of trajectories in each cluster for the car problem. Column: different state variables for each cluster. Row: state variable for different clusters.}
  \label{fig:carClusterTraj}
\end{figure}

The data is generated by uniformly sampling the parameter space.
Fig.~\ref{fig:carSNNbadPred} shows a few samples of the optimal trajectories.
Similar to the pendulum swingup problem, the constraint on $\theta_f$ makes it possible to reach the goal with different $\theta_f$.
The custom clustering is developed by inspection, whereby we first divide the dataset into three groups based on the final angle.
Then we find that for trajectories with the same $\theta_f$, the car can either go forward or backward to reach the origin, i.e. with positive or negative velocities.
This is illustrated in Fig.~\ref{fig:carClusterTraj}.
Hence, we divide the dataset into 6 clusters.
We note that the cluster sizes are bimodal and we use larger regression network for cluster with larger size.


\subsubsection{Trajectory tracking} Because this problem is not controllable at the origin, a stabilizing LQR controller may not be used at the trajectory endpoint.  Instead, we simply perform LQR rollout on the predicted trajectory, and stop when the end time is reached.  To determine success, we check if norm of final state error is within 0.5.

\subsubsection{Results and Discussion}
The data in Tab.~\ref{tab:summary} show similar trends to the pendulum problem, in particular, MoE yields lower validation error and higher rollout success rate than SNN. 
Moreover, the custom clustering outperforms k-Means which further outperforms SNN.
In Fig.~\ref{fig:carSNNbadPred} we show the predictions from SNN and MoE on a selected parameter as well as the optimal trajectories of its neighbors.
It is clearly shown that SNN may fail to predict $\theta_f$ correctly.

\begin{figure}
\centering
\begin{subfigure}[b]{0.475\columnwidth}
    \centering
    \includegraphics[width=\textwidth]{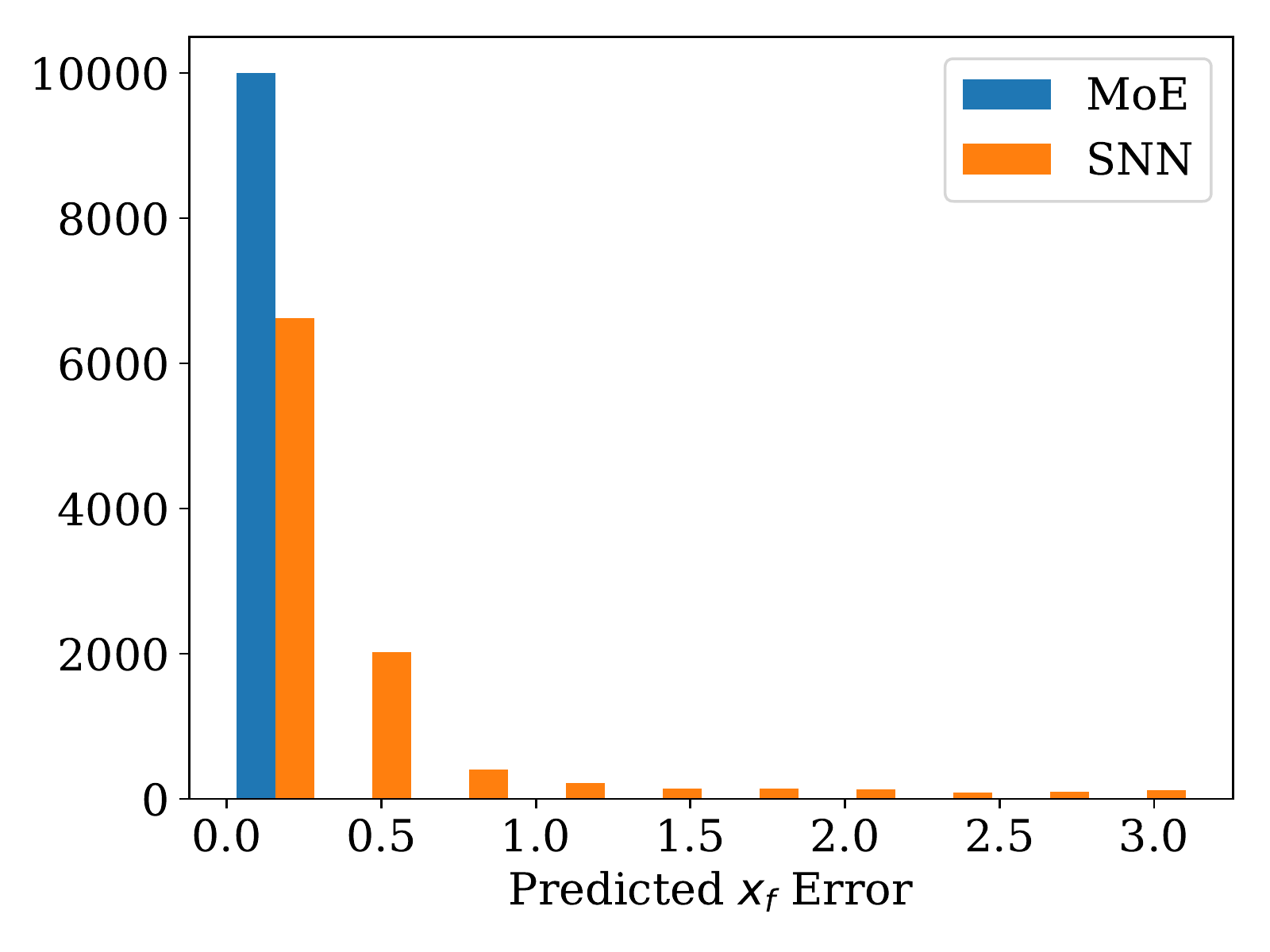}
    \caption[MoE Prediction]%
    {{\small Prediction error of $\bx_f$}}    
\end{subfigure}
\hfill
\begin{subfigure}[b]{0.475\columnwidth}  
    \centering 
    \includegraphics[width=\textwidth]{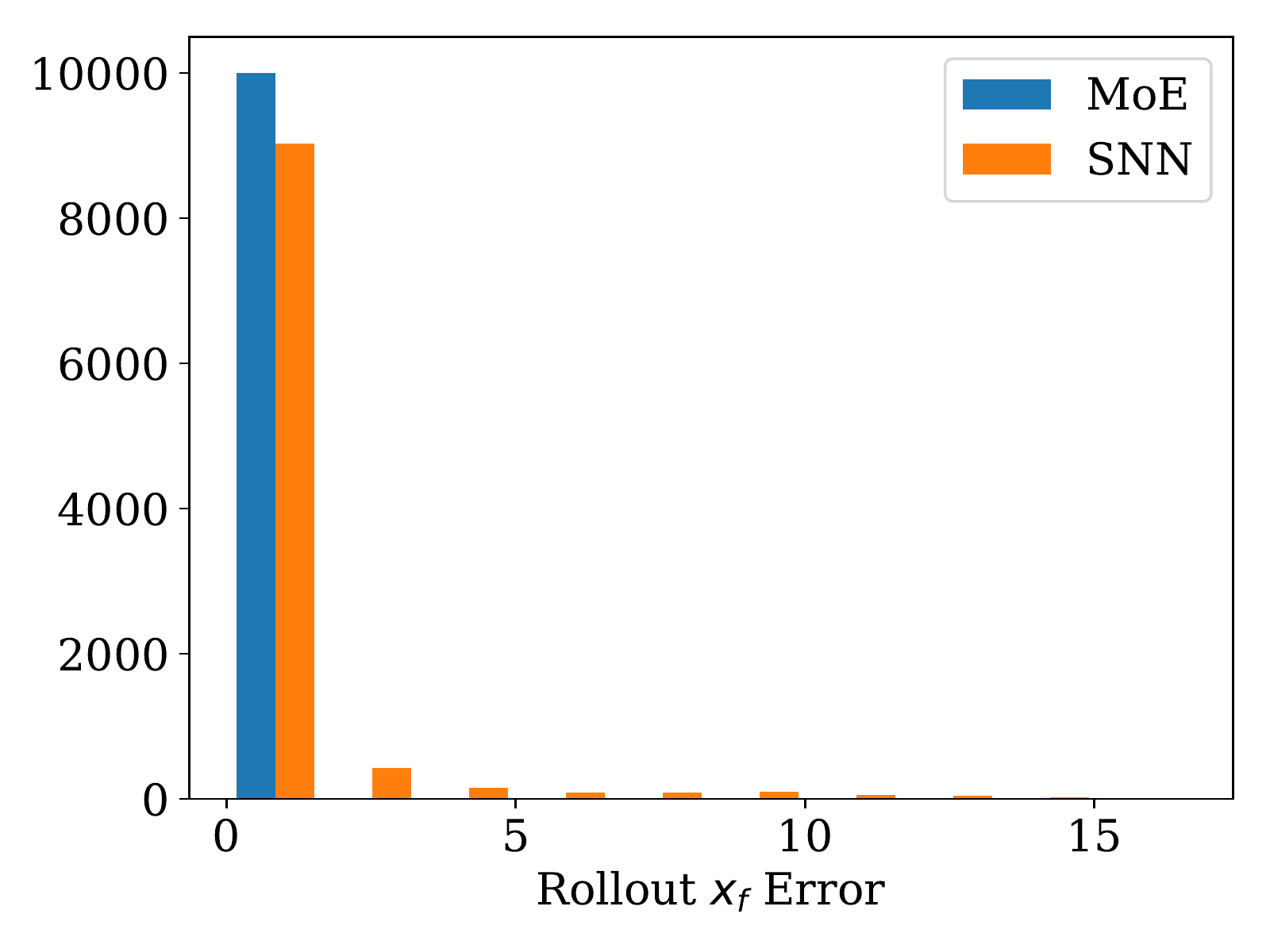}
    \caption[]%
    {{\small Tracking error of $\bx_f$}}    
\end{subfigure}
\caption
{\small Histograms of prediction and tracking results for MoE and SNN on the car problem.} 
\label{fig:carRolloutHist}
\end{figure}

The histogram in Fig.~\ref{fig:carRolloutHist}.a shows the norm of the error in predicted final state, indicating that SNN has higher prediction error. 
Fig.~\ref{fig:carRolloutHist}.b also show that paths predicted by SNN violates system dynamics more than MoE.
The reason why tracking error is actually much larger than predicted is that the predicted trajectory violates system dynamics, so path tracking diverges.

\subsection{Quadcopter with Collision Avoidance}
\subsubsection{Problem Setup}
The system has state $\bx=(x,y,z,v_x, v_y, v_z, \phi, \theta, \psi, p, q, r)\in\mathbb{R}^{12}$ and control $\bu\in\mathbb{R}^4$.
We refer \cite{Mellinger:2012ez} to the details.
The goal is to control the quadcopter from any equilibrium state with position within $[-10, 10]^3$ and all other states zero to the goal state $\bs{0}$.
The cost function is a weighted sum of time, control energy, and penalty on states, i.e. $J=w(t_f-t_0)+\int_{t_0}^{t_f}\bs{x}\T\bs{Q}\bs{x}+\bu\T\bs{R}\bu\dt$ with $w=10$, $\bs{Q}=\text{diag}(0, 0, 0, 1, 1, 1, 0.1, 0.1, 0.1, 1, 1, 1)$, $\bs{R}=\text{diag}(1, 1, 1, 1)$.

The quadcopter-obstacle case imposes additional path constraints on the state variables. The obstacle is a sphere with different position and radius, and obstacles are randomly placed in space with radius within $[1, 5]$.  We are interested in how the obstacles influence the trajectory.

\subsubsection{Data Generation and Training}

In the obstacle-free case, initial positions are sampled at random, and k-Means is used for clustering. 

\begin{figure}[tbp]
  \centering
  \includegraphics[width=\columnwidth]{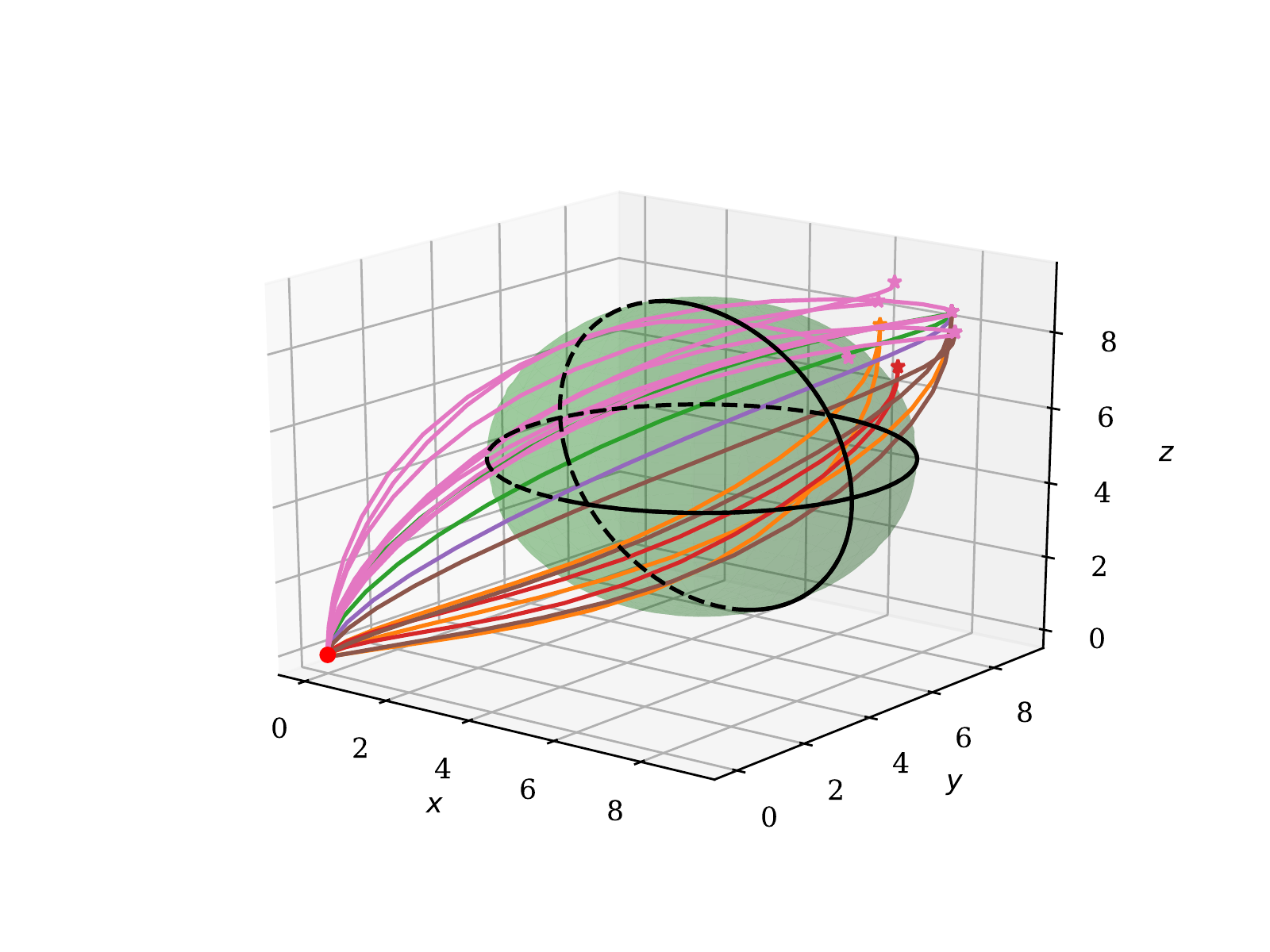}
  \caption{\small Trajectories of problems with parmameter close to the problem with sphere at (4, 4, 4), radius 3 and initial position (8, 8, 8). 
  Each color corresponds to trajectories from one cluster. It shows the trajectories can be quite different even for close problem parameters.}
  \label{fig:obsBunchTraj}
\end{figure}

\begin{figure}[tbp]
  \centering
  \includegraphics[width=\columnwidth]{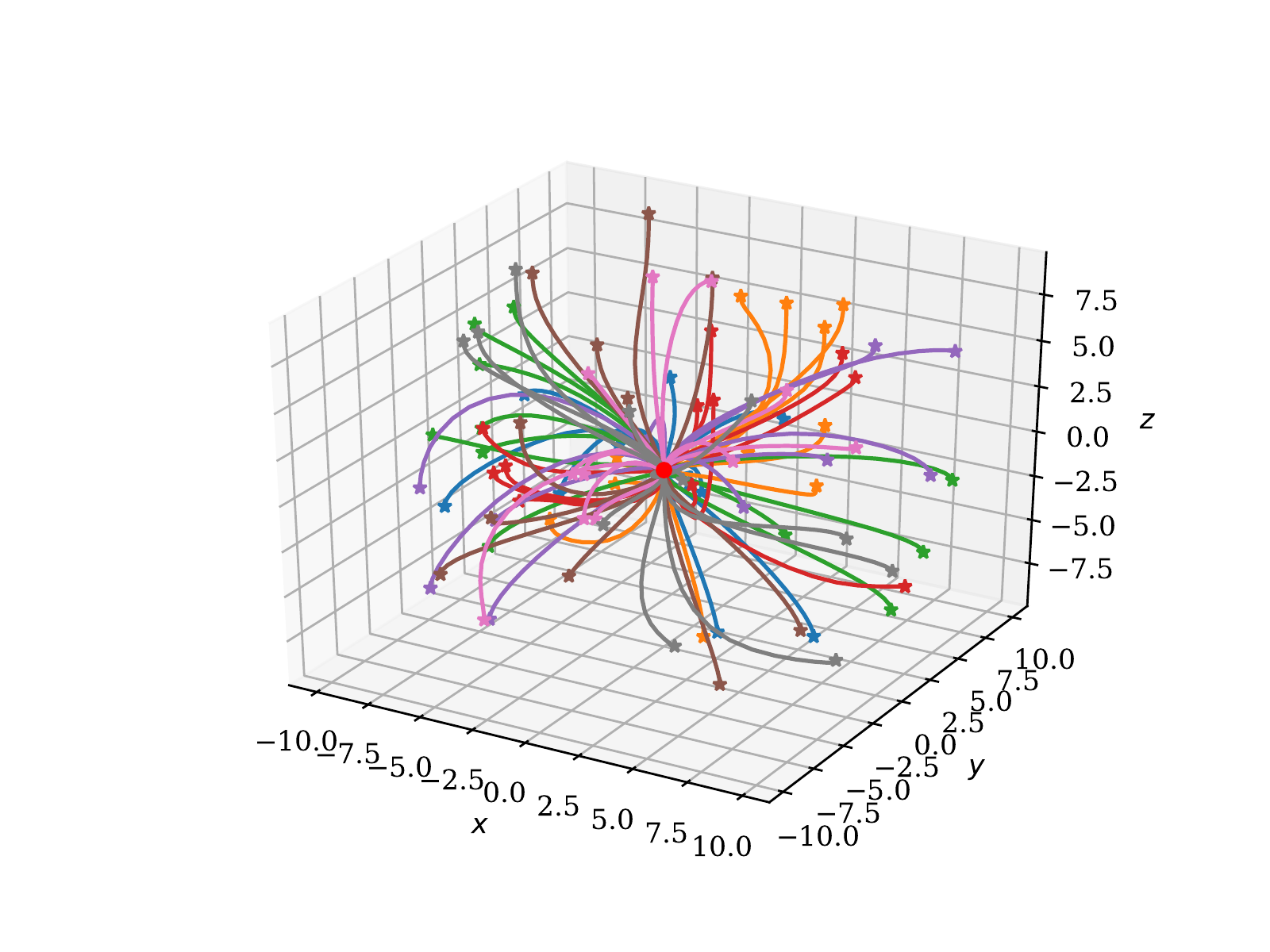}
  \caption{\small Samples of optimal trajectories for the quadcopter problem. Each color corresponds to each trajectory cluster.}
  \label{fig:obsClusTraj}
\end{figure}

The obstacle problem is more challenging because it has higher dimensionality in parameter space (7).
The OCP is also more challenging to solve due to the non-convex of obstacle avoidance constraint.  We want to focus on problem instances with significant obstacles, so our dataset only includes examples where the optimal collision-free trajectory would collide with an obstacle.  To generate this dataset, we collect  obstacle free trajectories and then sample obstacles that collide with the trajectory.  We then re-optimize for the sampled obstacles.
Samples of trajectories are shown in Fig.~\ref{fig:obsBunchTraj} and Fig.~\ref{fig:obsClusTraj}.

The discontinuity of the parameter-solution mapping in this problem is avoiding the obstacle from different directions outperforms others and vice versa.
One feature that describes how the obstacle-free trajectory is affected by the obstacles is the gradient of the active constraints with respect to state variables.
Since the obstacles are spheres, the gradient is essentially the vector from the center of the sphere to the point on surface where constraints are active.
Its direction clearly shows which direction the trajectory has to change for collision avoidance.
For trajectories that has more than one active constraints, we use the multipliers as weights and take the average.
In this way, a 3D vector is calculated for each trajectory and used as features to divide the problem space.
We divide the dataset into 8 groups based on the sign of each element of the 3D vector.

\subsubsection{Results and Discussion}

Results show that both SNN and MoE control the quadcopter to a stabilizable state in highly reliable fashion without obstacles. 
Hence, for validation we focus more on the amount of collision avoidance violation, i.e. $\min\{\|\bx_i-\bm{c}_o\|-r_o\}_{i=0}^N$ where $r_o$ and $\bm{c}_o$ are respectively the radius and center of the obstacle.

\begin{figure}[tbp]
  \centering
  \includegraphics[width=\columnwidth]{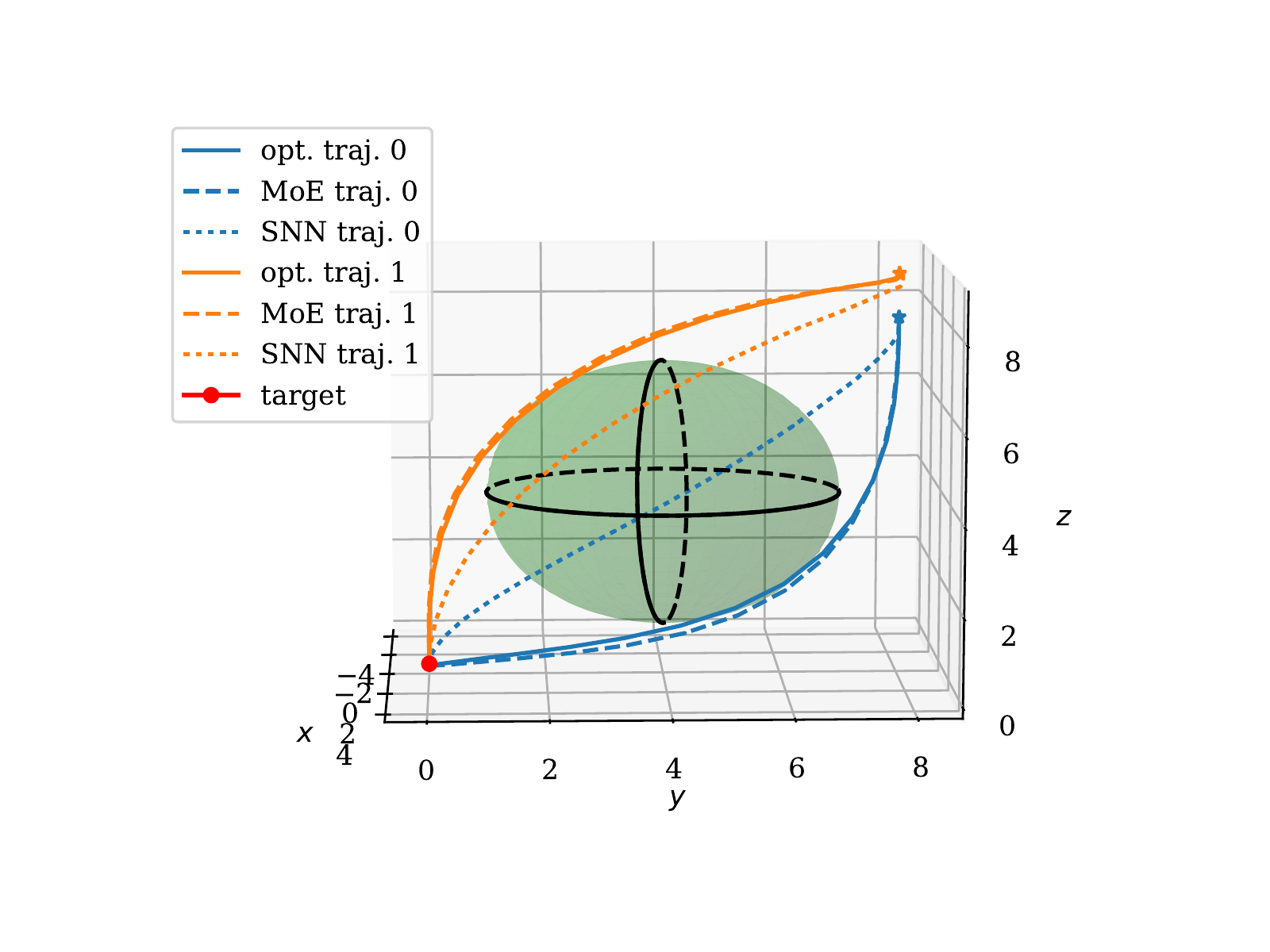}
  \caption{\small Optimal trajectories and prediction from SNN and MoE for two selected close states. 
  The green sphere is an obstacle centered at (0, 4, 4) with radius of 3.
  The solid, dashed and dotted lines are the optimal trajectories, prediction of MoE, and prediction of SNN, respectively.
  It shows SNN predicts a trajectory that violates obstacles avoidance constraints.}
  \label{fig:obsSNNbadMoEgood}
\end{figure}

\begin{figure}
\centering
    \includegraphics[width=\columnwidth]{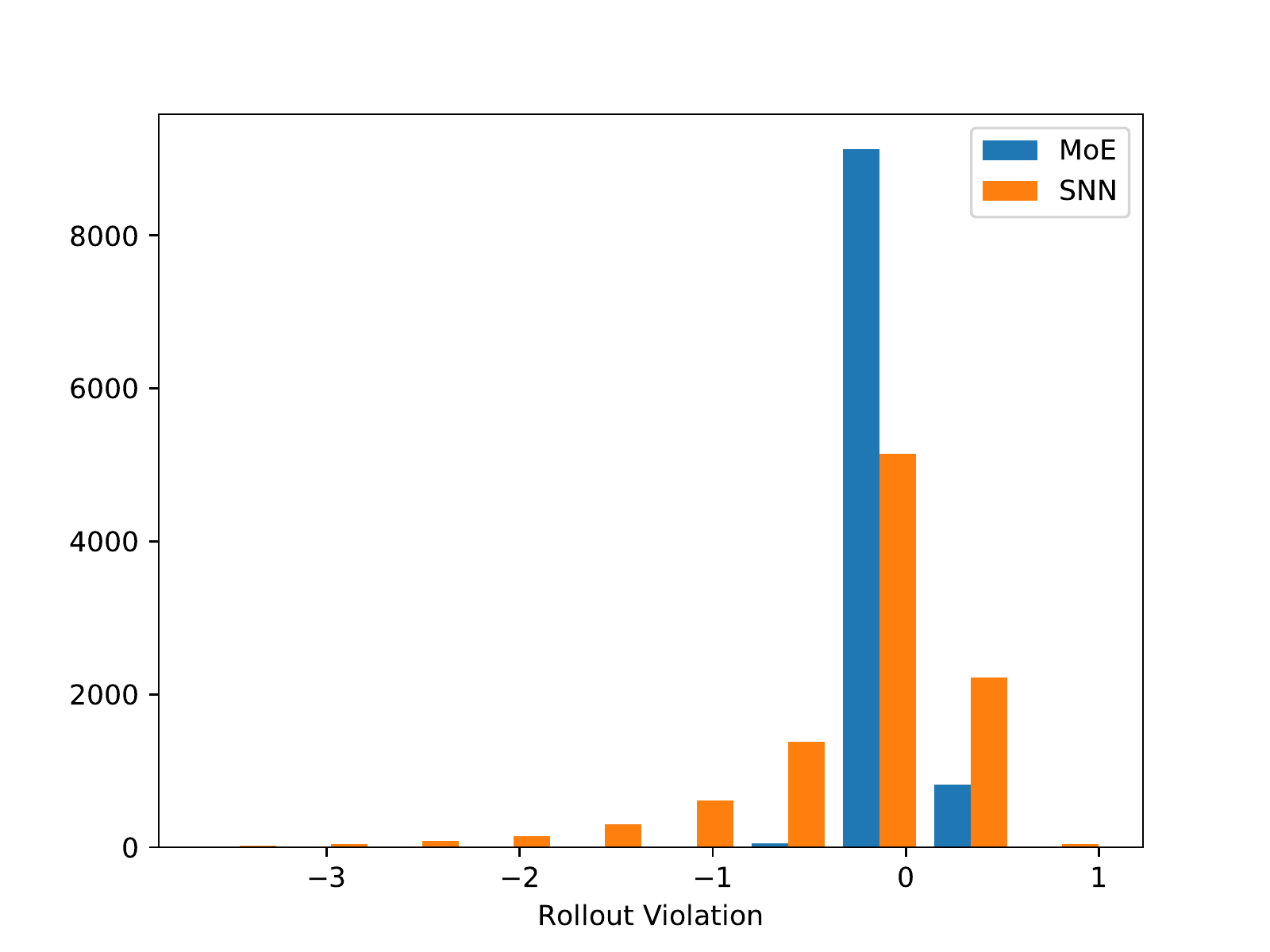}
    \caption
    {{\small Rollout constraint violation for quadcopter-obstacle.}}    
\label{fig:constrVio}
\end{figure}

With obstacles, MoE with custom cluster also significantly outperforms others.
A histogram of the constraint violation is shown in Fig.~\ref{fig:constrVio}, indicating that MoE yields much lower violation of constraints than SNN.
Fig.~\ref{fig:obsSNNbadMoEgood} shows examples of optimal trajectories and prediction from SNN and MoE.
As the initial state moves along $z$ direction, the optimal trajectories turns from going above to going below the obstacle.
SNN is unable to handle such discontinuity and predicts a trajectory that violates the constraints.
However, MoE is able to detect such discontinuity and predicts the corresponding trajectories.  It is important to note however that MoE still creates grazing collisions, so to successfully avoid an obstacle in practice, either a margin of error should be added to the modeled obstacle, or local collision avoidance should be added to the trajectory tracker.

\section{Conclusion} 
\label{sec:conclusion}

In this paper we demonstrate that optimal trajectories can be learned with high accuracy if we take into account the special structure of optimal control problems.
The mixture of experts model is designed such that each expert approximates a smooth region in the problem optimum map, and the classifier handles discontinuities without averaging.
It is important to train MoE with the correct clusters, and curiously, coupled training of the regressors and classifier tends to be detrimental to tracking performance.  We also argue that test error is not a good metric to judge learning models, but rather rollout success rate under trajectory tracking control is preferable.

Future work includes developing more sophisticated clustering algorithms that automatically find the best partition strategy.
For certain OCPs, differential flatness can be used such that the predicted trajectory satisfies dynamical constraints.
Further work also includes how to prove the stability of the predicted trajectories, and to scale up to handle larger problems, e.g., from sensor data or model uncertainties.

\section*{Acknowledgments}
This work is supported by NSF grant \#IIS-1816540.

\bibliographystyle{plainnat}
\bibliography{references}

\end{document}